\documentclass[10pt,twocolumn,letterpaper]{article}

\usepackage{cvpr}
\usepackage{times}
\usepackage{epsfig}
\usepackage{graphicx}
\usepackage{amsmath}
\usepackage{amssymb}

\newcommand{\tabincell}[2]{\begin{tabular}{@{}#1@{}}#2\end{tabular}}

\usepackage[pagebackref=true,colorlinks,bookmarks=false]{hyperref}

\cvprfinalcopy % *** Uncomment this line for the final submission

\def\cvprPaperID{8347} % *** Enter the CVPR Paper ID here

\ifcvprfinal\fi
\begin{document}
\setlength{\parskip}{0pt}
\setlength{\abovedisplayskip}{4pt}
\setlength{\belowdisplayskip}{4pt}

%%%%%%%%% TITLE
\title{P$^2$GNet: Pose-Guided Point Cloud Generating Networks for 6-DoF Object Pose Estimation}

\author{Peiyu Yu, Yongming Rao, Jiwen Lu, Jie Zhou\\
Department of Automation, Tsinghua University, China\\
State Key Lab of Intelligent Technologies and Systems, China\\
Beijing National Research Center for Information Science and Technology, China \\
{\tt\small yupy16@mails.tsinghua.edu.cn; raoyongming95@gmail.com; \{lujiwen, jzhou\}@tsinghua.edu.cn}
}

\maketitle
%\thispagestyle{empty}

%%%%%%%%% ABSTRACT
\begin{abstract}
    % Six Degree-of-Freedom (6-DoF) object pose estimation, the problem of estimating the 3D translation and 3D rotation of objects from partial observations, lies at the core of many vision and robotics applications. In this work, we propose \underline{P}ose-Guided \underline{P}oint Cloud \underline{G}enerating Networks (P$^2$GNet), a novel learning-based approach. Unlike existing pose estimation methods, P$^2$GNet aims to address the ambiguity of the object appearance caused by single-view input with rich structure knowledge provided by object point cloud, rather than exhausting the limited appearance information from RGB-D image. It features an end-to-end estimation-by-generation workflow, where the networks are jointly trained by pose estimation task and pose-guided point cloud generating tasks, enforcing the integration of object prior into the existing state-of-the-art 6-DoF pose estimation system. Experiments on two commonly used benchmarks for 6D pose estimation, YCB-Video dataset and LineMOD dataset, demonstrate that P$^2$GNet outperforms the competing methods by a large margin and shows marked robustness towards heavy occlusion, while achieving real-time inference.
   Humans are able to perform fast and accurate object pose estimation even under severe occlusion by exploiting learned object model priors from everyday life. However, most recently proposed pose estimation algorithms neglect to utilize the information of object models, often end up with limited accuracy, and tend to fall short in cluttered scenes. In this paper, we present a novel learning-based model, \underline{P}ose-Guided \underline{P}oint Cloud \underline{G}enerating Networks for 6D Object Pose Estimation (P$^2$GNet), designed to effectively exploit object model priors to facilitate 6D object pose estimation. We achieve this with an end-to-end estimation-by-generation workflow that combines the appearance information from the RGB-D image and the structure knowledge from object point cloud to enable accurate and robust pose estimation. Experiments on two commonly used benchmarks for 6D pose estimation, YCB-Video dataset and LineMOD dataset, demonstrate that P$^2$GNet outperforms the state-of-the-art method by a large margin and shows marked robustness towards heavy occlusion, while achieving real-time inference.
\end{abstract}

%%%%%%%%% BODY TEXT
% \vspace{-0.5cm}
\section{Introduction}
6-DoF object pose estimation lies at the core of a wide range of applications including robotic manipulation \cite{Zhu2014Single, Collet2011MOPED}, augmented reality \cite{Marchand2016AR}, navigation and autonomous driving  \cite{Chen2016multiview3d, Geiger2012KITTI, Xu2018Pointfusion}. Ideally, a handy 6D object pose estimation system should deal with objects of varying shape and texture, achieving marked robustness towards heavy occlusion, sensor noise, and changing lighting conditions, while meeting the speed requirement of real-time tasks. % The advent of cheap RGB-D sensors has enabled methods that infer poses of low-textured objects even in poorly-lighted environments more accurately than RGB-only methods.

% Prior to the recently emerged data-driven methods, traditional approaches address this problem by extracting features from RGB-D data and performing correspondence grouping and hypothesis verification \cite{inproceedings, article1, 7845723}.
\begin{figure}[htbp]
\begin{center}
  \includegraphics[width=0.45\textwidth]{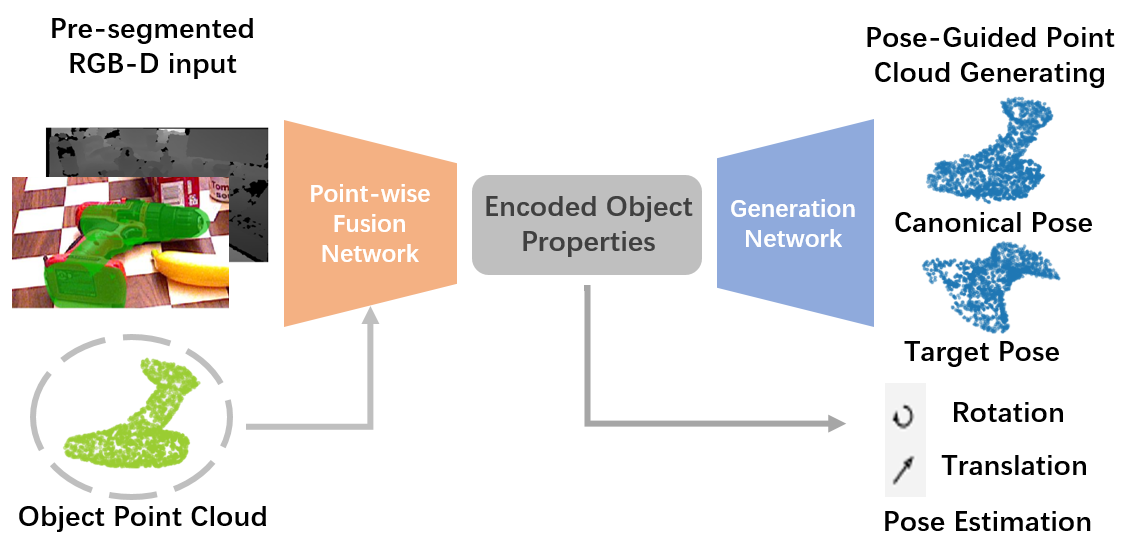}
\end{center}
  % \vspace{-5pt}
  \caption{An example of our pose-guided point cloud generation scheme. Our model takes object point cloud in canonical pose and RGB-D image as input, and generates point clouds in both canonical pose and target pose to faciliate pose estimation.}
  \label{fig_example}
  % \vspace{-10pt}
\end{figure}
For many years the main focus in the field of 6D pose estimation of objects has been limited to geometric feature/template-based methods. Based on the pioneering work of \cite{Hinterstoisser2011linemod}, practical and robust solutions have been designed to handle objects with poor textures \cite{Hinterstoisser2012ADD, Rios2013Discriminatively, Brachmann2014L6D, Kehl2016Local, Hinterstoisser2016PPF,  Tejani2014Latent, Brachmann2016Uncertainty, Buch2017Rotational}. For most of these systems, the key to success is the use of either hand-crafted templates, or feature representations extracted from data. These systems run typically a two-stage pipeline: a) putative feature matching, b) geometric verification of the matched features. Nevertheless, the techniques mentioned above have in our view several fundamental shortcomings. Firstly, there are common problems of these feature/template-based techniques, such as not being robust to clutter and occlusions as well as changing lighting conditions. Secondly,  scalability to large numbers of objects (i.e., either object instances or object categories) is an open challenge to these techniques, due to the growing number of required features/templates. Thirdly, the reliance on hand-crafted features, fixed matching procedures, and particularly on verification procedures have made it difficult for them to satisfy the requirements of accurate pose estimation and fast inference simultaneously.

Recent success in visual recognition has inspired a novel family of data-driven methods using deep networks to surmount the limitations (especially the scalability) raised above \cite{Yi2018DeepIM, Sundermeyer2018Implicit, Xiang2018posecnn, Li2018MCN, Tremblay2018Deep, Tekin2017RealTime, Schwarz2015RGB}. PoseCNN \cite{Xiang2018posecnn} and MCN \cite{Li2018MCN} have achieved decent accuracy on YCB-Video dataset using single-view and multi-view inputs respectively. However, similar to the prior methods, these methods still require elaborate post-processing refinement steps to fully utilize the depth information, such as the computationally expensive Iterative Closest Point (ICP) \cite{Fisher01projectiveicp} procedure in PoseCNN and the multi-view hypothesis verification scheme in MCN, preventing them from satisfying the requirements of real-time inference. Inspired by the recent success in 3D object recognition that combines the information from both RGB camera and depth sensor, such as Frustrum PointNet \cite{Qi2018frustum} and PointFusion \cite{Xu2018Pointfusion}, DenseFusion \cite{Wang2019DenseFusion} propose to better exploit the complementary nature of color and depth information from RGB-D data with an end-to-end deep model, and has achieved promising performance and the capacity of real-time inference. Still, method using only single-view RGB-D data is inherently incapable of addressing the ambiguity of object appearance \cite{Li2018MCN}, which can potentially harm the performance.

Unlike automatic systems that are susceptible to environment, sensor noise and object appearance, humans are capable of performing decent and fast object pose estimation, even under adversarial conditions (i.e., heavy occlusion and changing ambient lighting). Our assumption is that humans exploit learned object model priors from everyday life, and therefore can readily and accurately infer the object pose. Hence, we attempt to explore how the prior knowledge of object can be integrated into state-of-the-art single-view 6D object pose estimation systems, so that our method can 1) perform more accurate pose estimation without prohibitively expensive post-hoc steps, 2) achieve marked robustness towards heavy occlusion, and 3) satisfy the speed requirement of real-time tasks.

In this work we propose an end-to-end deep learning approach for estimating 6-DoF poses of known objects from RGB-D inputs and object point clouds. The core of our approach is to sufficiently explore the intrinsic properties of object point clouds by using \emph{pose-guided point cloud generating} as proxy task. To be specific, our method takes object point cloud in canonical pose and RGB-D image as inputs, generating object point cloud in the pose corresponding to the RGB-D input to facilitate the object pose estimation. The estimation-by-generation scheme enables our method to explicitly encode both the appearance and 3D geometric information, which is essential to address the ambiguity of object appearance and occlusion in cluttered scenes. As demonstrated by our experiments, by effectively exploiting object model priors, our method combines RGB-D data and 3D point clouds representation in a more effective manner at per-point level, opposing prior work which only uses 2D or limited 3D information to compute dense features \cite{Wang2019DenseFusion, Qi2018frustum}. We evaluate our method on two popular benchmarks for 6D pose estimation, YCB-Video dataset \cite{Xiang2018posecnn} and LineMOD dataset \cite{Hinterstoisser2012ADD}. Experiments show that our method alone (i.e., without any post-processing steps) outperforms DenseFusion without iterative refinement \cite{Wang2019DenseFusion} significantly by 10.0\% in ADD pose accuracy \cite{Hinterstoisser2012ADD}, and even outperforms DenseFusion with iterative refinement by 1.9\% on LineMOD dataset. On YCB-Video dataset, P$^2$GNet shows marked improvement of 2.3\% and 0.8\% in terms of ADD-S$<$2cm metric comparing with the state-of-the-art DenseFusion without and with iterative refinement respectively. Moreover, we demonstrate that P$^2$GNet can also be augmented by iterative refinement, achieveing 99.9\% accuracy in terms of ADD-S$<$2cm metric and 93.9\% in terms of AUC metric \cite{Xiang2018posecnn} on YCB-Video dataset. It is fair to say that our method further improves the state-of-the-art system, and to our knowledge surpasses all the competing network-based methods without laborious tuning, while maintaining the efficiency in inference time.

We summarize our contributions in this work as follows:
% \vspace{-15pt}
\begin{itemize}
    \item We propose an effective way to integrate the prior knowledge of object into state-of-the-art single-view 6D object pose estimation systems;
    \item We demonstrate that the proposed method significantly outperforms the competing methods, approaching and exceeding SOTA without post-hoc steps on two benchmark datasets respectively;
    \item We show that our method is capable of robust pose estimation under heavy occlusion, efficient enough for real-time inference ($\sim$1.5ms per object instance) and extendable for better performance (ADD: 97.4\% on LineMOD and ADD-S$<$2cm: 99.9\%, AUC: 93.9\% on YCB-Video dataset with iterative refinement).
\end{itemize}

%------------------------------------------------------------------------
\section{Related Work}
\textbf{3D representations.} Most deep neural networks for 3D inputs model the 3D space as pre-partitioned regular voxels, and extend 2D convolution to voxel-based 3D convolution, such as \cite{Brock2016GDVM, Dai2017scannet, Wu20163DGAN}. The main problem of voxel-based methods is the increasing spatial resolution which leads to fast growth of neural-network size and computational cost. Following the voxel-based methods, a family of methods using elaborately designed sampling strategy such as octree-based \cite{Riegler2017octnet} and kd-tree-based \cite{klokov2017kdnet} neural networks partly overcome the challenge. Recently, neural networks based on purely 3D point representations (i.e., 3D coordinates) \cite{achlioptas20173dad, Qi2016PointNet, Qi2017pointnet++, Yang2017FoldingNet} have been shown to work quite efficiently, while achieving promising performance on several 3D recognition tasks. Point-based neural networks significantly reduce the overhead of converting point clouds into other data formats (such as trees and voxels), circumventing the potential information loss due to the conversion.

\begin{figure}[htbp]
\begin{center}
  \includegraphics[width=0.45\textwidth]{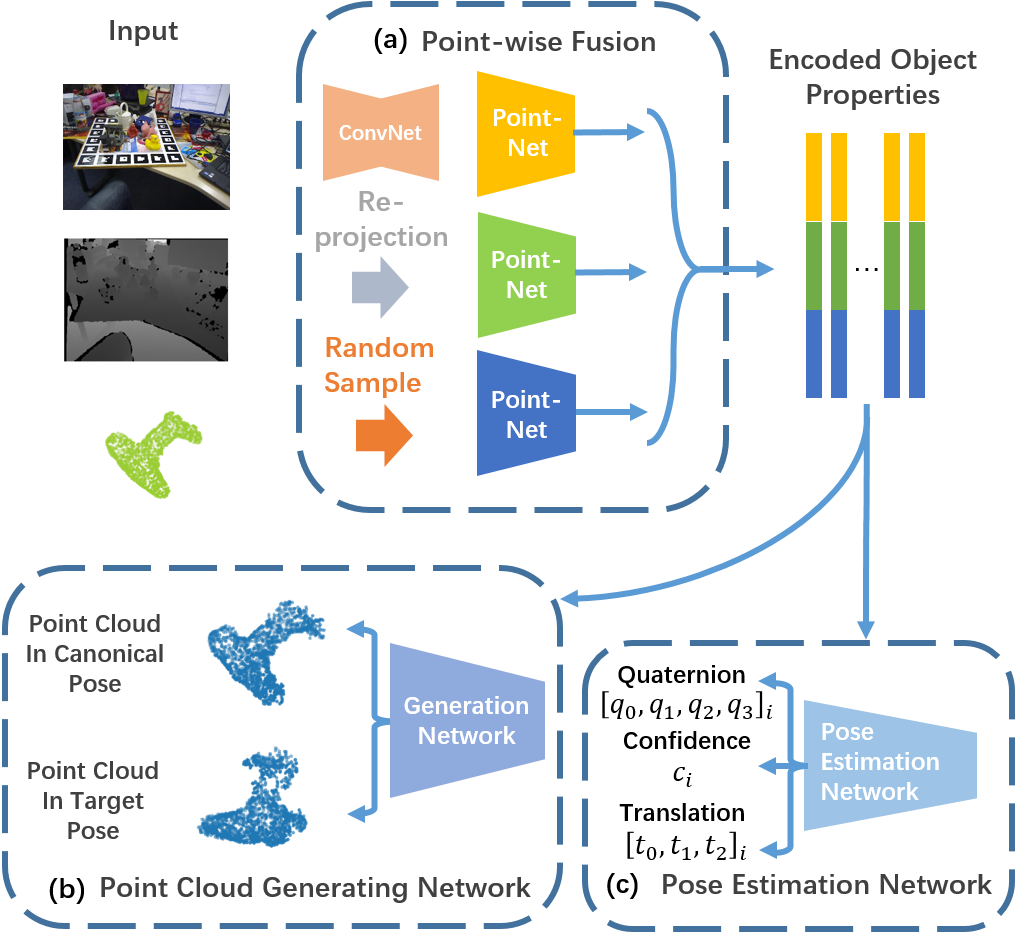}
\end{center}
    \caption{Our model for 6D object pose estimation has three components: (a) a point-wise fusion network, (b) a point cloud generating network, and (c) a pose estimation network. Best viewed in color.}
  \label{fig_model}
  % \vspace{-15pt}
\end{figure}

Prior to this work, Frustrum PointNets \cite{Qi2018frustum} and VoxelNet \cite{Zhou2017voxelnet} use a PointNet-like \cite{Qi2016PointNet} structure and achieve state-of-the-art performances on the KITTI benchmark \cite{Geiger2012KITTI} in tasks regarding on autonomous driving, demonstrating the effectiveness of this architecture. However, urban driving scenarios differ from those of indoor pose estimation since adversarial conditions such as heavy occlusion and sensor noises are more common and severe in the latter ones. In this work, we borrow the idea of directly processing 3D point representations to achieve real-time inference as well as decent accuracy under heavy occlusion.

% \textbf{Pose from RGB images.} Classical methods rely on detecting and matching keypoints with known object models [\cite{Aubry2014Seeing}, \cite{article}, \cite{ferrari}, \cite{roth}, \cite{6907430}]. Newer methods address the challenge by learning to predict the 2D keypoints [\cite{inproceedings}, \cite{Pavlakos20176}, \cite{latentk3y}, \cite{Tekin2017Real}, \cite{Tremblay2018Deep}] and solve the poses by PnP [\cite{pnp}]. Though prevail in speed-demanding tasks, these methods become unreliable given low-texture or low-resolution inputs. Other methods propose to directly estimate objects pose from images using CNN-based architectures [\cite{Schwarz2015RGB}, \cite{Tulsiani2015Viewpoints}]. Many such methods focus on orientation estimation: \citet{Yu2015Data}, \citet{Yu2016Subcategory} learns a viewpoint-aware pose estimator by clustering 3D features from object models. \citet{Mousavian20163D} predicts 3D object parameters and recovers poses by single-view geometry constraints. \citet{Sundermeyer2018Implicit} implicitly encode orientation in a latent space and in test time find the best match in a code book as the orientation prediction. However, pose estimation in 3D remains a challenge for the lack of depth information. Our method leverages both image and 3D data to estimate object poses in 3D in an end-to-end architecture by domain adaptation.

\textbf{Pose from RGB/RGB-D images.} For many years, approaches for 6D pose estimation focus on constructing appropriate geometric features/templates from the input RGB/RGB-D images and perform correspondence grouping and hypothesis verification \cite{Hinterstoisser2011linemod, Hinterstoisser2012ADD, Rios2013Discriminatively, Hinterstoisser2016PPF, Buch2017Rotational}. These features/templates methods are either hard coded \cite{Hinterstoisser2011linemod}, or requiring extra tuning when applied to different objects \cite{Hinterstoisser2012ADD, Rios2013Discriminatively, Hinterstoisser2016PPF}. \cite{Brachmann2014L6D, Brachmann2016Uncertainty, Tejani2014Latent, Wohlhart2015Descriptors} as alternatives, propose methods optimizing surrogate objectives, and provide practical and robust solutions. In a recent benchmark BOP \cite{Hodan2018BOP} for this task , where no or few real training data is available but only CAD models for rendering synthetic training data, i.e.,  learning-based methods have to handle a severe domain gap between training and testing, purely geometric approaches (based on \cite{Hinterstoisser2016PPF}) significantly outperform learning-based approaches. However, as we demonstrate in the introduction part, these methods fail to simultaneously achieve good performance and fast inference speed, while scalability remains a critical issue.

Newer network-based methods such as PoseCNN \cite{Xiang2018posecnn} and MCN \cite{Li2018MCN} directly estimates 6D poses from image data using a CNN-based architecture, while they rely on expensive post-processing steps to make full use of depth input. Inspired by the recent success in 3D object recognition, DenseFusion \cite{Wang2019DenseFusion} propose to better exploit the complementary nature of color and depth information with an end-to-end network, and has achieved promising performance and the capacity of real-time inference. Nevertheless, from our view, methods using single-view RGB-D data can't fully capture the 3D object shape which is essential in 6D pose estimation, since the inputs lack the deterministic information to address the ambiguity of object appearance. We show that our method tackle this challenge by effectively learning the 3D object information, therefore outperforming most competing methods and achieving fast inference speed in the mean time.

% \textbf{Pose from point cloud.}

Our method is most related to DenseFusion \cite{Wang2019DenseFusion}, in which researchers perform local feature fusion between color and depth information using a heterogeneous architecture. By introducing object point clouds as the object prior, and performing pose-guided point cloud generating task, our novel estimation-by-generation scheme surpasses DenseFusion’s estimation-by-fusion method by a large margin. Since our method is a step forward from DenseFusion, we demonstrate that our method is also suitable for the iterative refinement method \cite{Wang2019DenseFusion}, which leads to extra improvements in pose estimation, while maintaining the fast inference speed.

\section{Approach}

Accurately estimating the pose of a known object in adversarial conditions (e.g. heavy occlusion, poor lighting, \textit{etc.}) is only possible when deterministic geometric and appearance information are accessible. Therefore, pose estimation in 3D space using single-view RGB-D images remains a challenge since the inputs lack the critical 3D shape information to address the ambiguity of object appearance. To overcome this challenge, we propose to leverage the intrinsic properties of object models to guide pose estimation. More specifically, a novel pose-guided point cloud generating framework is designed, where we fuse the appearance information from RGB-D image and the structure knowledge from object point cloud to predict the 6D object pose. The pipeline of our method is presented in Figure~\ref{fig_model}.

In the following sections, we first introduce our design on pose-guided point cloud generating networks, which extract the knowledge of object models from given point cloud and incorporate this knowledge with RGB-D input to facilitate pose estimation (Sec.~\ref{3.1}). Then, we show the learning objectives of our method (Sec.~\ref{3.2}). Finally, we present the training paradigm and other technical details of our method (Sec.~\ref{3.3}),

\subsection{Pose-Guided Point Cloud Generating Networks}
\label{3.1}
\begin{figure*}[htbp]
  \begin{center}
  \includegraphics[width=\textwidth]{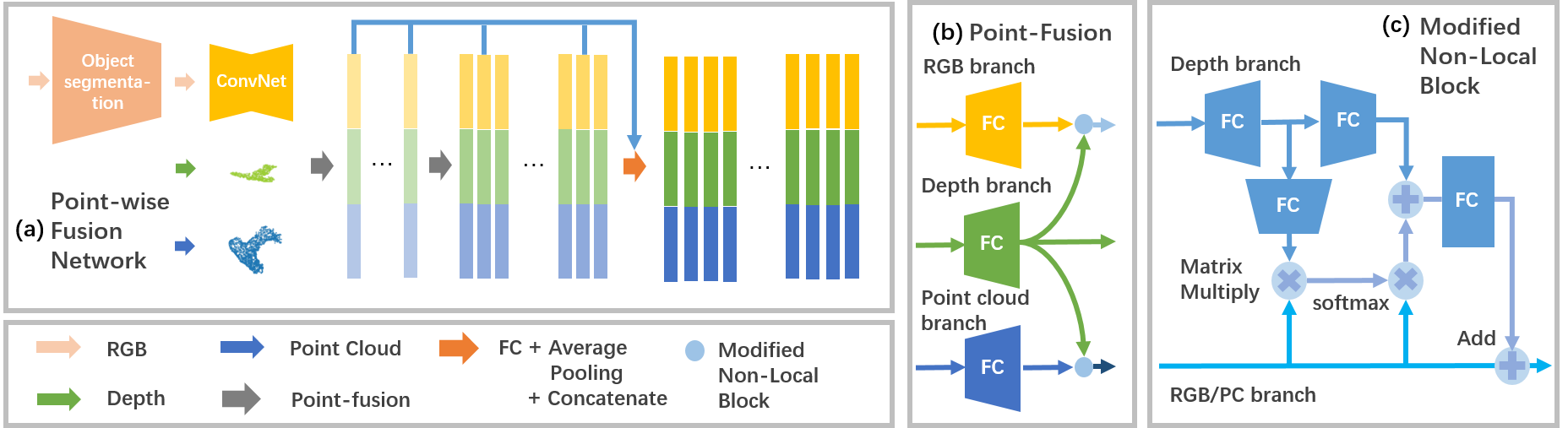}
  \end{center}
  % \vspace{-5pt}
    \caption{Detailed Point-wise Fusion architecture: (a) Point-wise Fusion pipeline, (b) point-fusion block and (c) modified non-local block.}
  \label{fig_point_fusion}
  % % \vspace{-10pt}
\end{figure*}
The proposed pose-guided point cloud generating networks (P$^2$GNet) are  comprised of three basic components: (a) a point-wise fusion network that fuses the information from both RGB-D image and object point cloud in canonical pose to encode the intrinsic object properties (Sec.~\ref{3.1.1}), (b) a point cloud generating network that utilizes the encoded feature from fusion network and performs pose-guided point cloud generation (Sec.~\ref{3.1.2}), and (c) a pose estimation network which takes the fused feature from point-wise fusion network to estimate the 3D translation and 3D rotation (Sec.~\ref{3.1.4}). Details of our architecture are described below.

% \vspace{-10pt}
\subsubsection{Point-wise Fusion Network}
\label{3.1.1}

The first component of our model (Figure~\ref{fig_model}a) takes both the RGB-D image and the object point cloud as inputs, generating fused feature that represents the encoded object properties. More specifically, to emphasize the appearance information of a object, we use the pre-trained segmentation architecture proposed by \cite{Xiang2018posecnn} to segment the objects of interest in the RGB-D images, and project the segmented depth pixels into 3D representations by employing the known camera intrinsic matrix before feeding them into the rest of the P$^2$GNet pipeline. At the beginning of point-wise fusion network, we use a CNN-based embedding network to obtain the deep representation of the object from RGB image, after which we randomly sample $N$ data points from each data branch (i.e., deep representation, re-projected depth values and object point cloud). The goal of this step is to form a sparse representation of each data source, thereby enabling the use of the PointNet-like architecture (i.e., MLP followed by a pooling layer) \cite{Qi2016PointNet} to realize fast inference.

The main idea of this network is to extract intrinsic object properties from different data sources, while discarding properties that are non-essential for the task of 6D pose estimation, such as object texture, lighting conditions and sensor noises. A simple but effective design is to follow DenseFusion \cite{Wang2019DenseFusion}, using three branches of MLP that separately process deep representation of RGB image, re-projected depth value and sampled object point cloud, and fuse the features by concatenation to generate a dense per-point feature. Since the pose-guided point cloud generating task serves as a strong per-point supervision, it seems natural and justified to do so. However, we have to point out that matching each point in the object point clouds with exactly the corresponding points from RGB-D images is rather difficult, and the mismatched object point cloud can therefore be a potential perturbation degrading the model performance, as we demonstrate in the ablative experiments (see Sec~\ref{4.3}). In order to address the problem, we propose the point-fusion block, in which each FC layer is followed by a modified non-local block \cite{Wang2018NonLocal}. We find that using point-fusion blocks significantly improves the fusion quality, compared with a naive solution using the plain MLPs and choosing the corresponding point for each point from depth image randomly. Detailed architecture of the point-fusion block and the network architecture are illustrated in Figure~\ref{fig_point_fusion}.

We set the output dimension of all of the three branches to $d_{emb}$ and concatenate the corresponding features, by which feature of each point becomes a $d_{emb} \times 3$-dim vector representing the appearance (from the RGB image) and geometric (from the depth image and the object point cloud) information of the input at the corresponding location. Then, the concatenated features are processed by a PointNet-like architecture as shown in Figure~\ref{fig_model}a. Here we obtain a fixed-size global feature vector by performing global average pooling on the processed features as in \cite{Wang2019DenseFusion}, with which we enrich each point-feature to provide a global context. We use squeeze-and-excitation block \cite{Hu2017Se} at the end of point-wise fusion network to increase its sensitivity to informative features.
% \vspace{-10pt}
\subsubsection{Point Cloud Generating Network}
\label{3.1.2}
We expect that the encoded feature produced by the Point-wise Fusion Network contains the complete 3D information about the object, and thus can address the ambiguity of the appearance caused by single-view inputs and more precisely predict the 6D pose. Here we introduce point cloud generating tasks as the supervision to achieve more effective feature fusion and facilitate the extraction of intrinsic properties of object models. To do so, we use two decoder branches (Figure~\ref{fig_model}b) to separately generate object point cloud in canonical pose and target pose. We generate the object point cloud in target pose to ensure that the feature generated by point-wise fusion network successfully encodes the intrinsic properties of object model. We generate the object point cloud in canonical pose to stabilize the training process and enforce that the encoded feature contains the complete 3D information of object model.

Intuitively, these branches compel the point-wise fusion network to properly learn the information from point cloud models, and explicitly infer the pose information embedded in the fused feature. By fully exploiting the information from point cloud representation, the network can achieve extra robustness and superior accuracy even in heavy occluded scenes, as we show in Sec.~\ref{4.3} and Sec.~\ref{4.4}. Compared with previous works that directly use pose estimation error to supervise models, the proposed estimation-by-generation method can provide stronger and more detailed (i.e., per-point) supervision to help the networks better understand the pose as well as the structure knowledge of input object. In our experiments, we implement a variant of the folding-based decoder \cite{Yang2017FoldingNet} that utilizes Gaussian sampled 3D grids to obtain decent generation results.

%\subsubsection{Fusion Network}
%\label{3.1.3}
%Encoder path that generates dense encoded feature can perform well on pose-guided point cloud generation, but tends to have an inferior performance on 6D pose estimation, considering that the dense, per-point supervision resulted from point cloud generation is much stronger than that from 6D pose estimation. Encoder path that generates sparse feature can readily find a balance between 6D pose estimation and point cloud generation tasks, whereas the generated feature may contain misleading parts resulted from non-target objects or the background due to occasional occlusion and segmentation errors, and therefore, would potentially degrade the model performance on estimation. Hence, we adopt a fusion network combining features obtained from previous step to make full use of the distilled intrinsic properties of object models.

%The main idea of the fusion network (Figure~\ref{fig_model}c) is to perform feature fusion not only over the final feature representation but also over the intermediate feature representation, so that the intermediate fused feature can provide the guidance information for feature selection. We adopt the squeeze-and-excitation operation [\cite{se}] to further improve the effectiveness of intermediate feature fusion. To reduce the computational complexity of 6D pose estimation, the fused feature is compressed to a fixed-size $1 \times d_{bottleneck}$ sparse feature vector.

% \vspace{-10pt}
\subsubsection{Pose Estimation Network}
% \vspace{-2pt}
\label{3.1.4}
We feed the features produced by point-wise fusion network into a final network (Figure~\ref{fig_model}c) that predicts the object’s 6D pose. % Here we also employ the architecture of folding-based decoder [\cite{folding}] and use separate branches to perform 3D translation estimation and 3D rotation regression. The last fully connected layer of the decoder branch has dimension $3 \times n$ or $4 \times n$, where $n$ is the number of object classes. For each class, the last FC layer outputs a 3D translation represented by a 3D vector or a 3D rotation represented by a quaternion respectively. We also integrate the same iterative refinement network architecture within our model as in DenseFusion [\cite{Wang2019DenseFusion}].
Note that here we predict the 6D pose from the features instead of the generated point clouds to reduce the computational cost. By doing so, the proposed generation networks are only used during training to provide supervision signal, which can largely reduce redundant computation for fast inference. Surprisingly, we find that directly predicting the 6d pose from encoded feature can achieve even better performance compared to predicting pose from generated point cloud in our experiments. We assume that the encoded feature produced by our framework offers better global information and sufficient detail information for pose estimation benefiting from strong supervision provided by generation network.

In our experiments, we use a similar pose estimation network as in DenseFusion \cite{Wang2019DenseFusion} to make fair comparisons. The only differences lie in the following three points: a) the input channel number is slightly different, which is inevitable since we introduce the object point cloud as an additional input; b) instead of following the principle of “object per output branch”, whereby each object class is associated with an output stream (i.e., parallel FC layers or MLPs) connected to a shared feature basis, we use a single MLP as object point cloud has provided sufficient class information and we want to disentangle the network size from the amount of objects; c) instead of using sigmoid activation function to produce the per-point confidence score, we use softmax activation function to normalize the $N$ confidence scores so that we can simplify the learning objective (see Sec~\ref{3.2}), discarding the confidence regularization term as well as its hyper-parameter proposed by DenseFusion \cite{Wang2019DenseFusion}.

% % \vspace{-0.3cm}
\begin{figure*}[htbp]
\begin{center}
  \includegraphics[width=\textwidth]{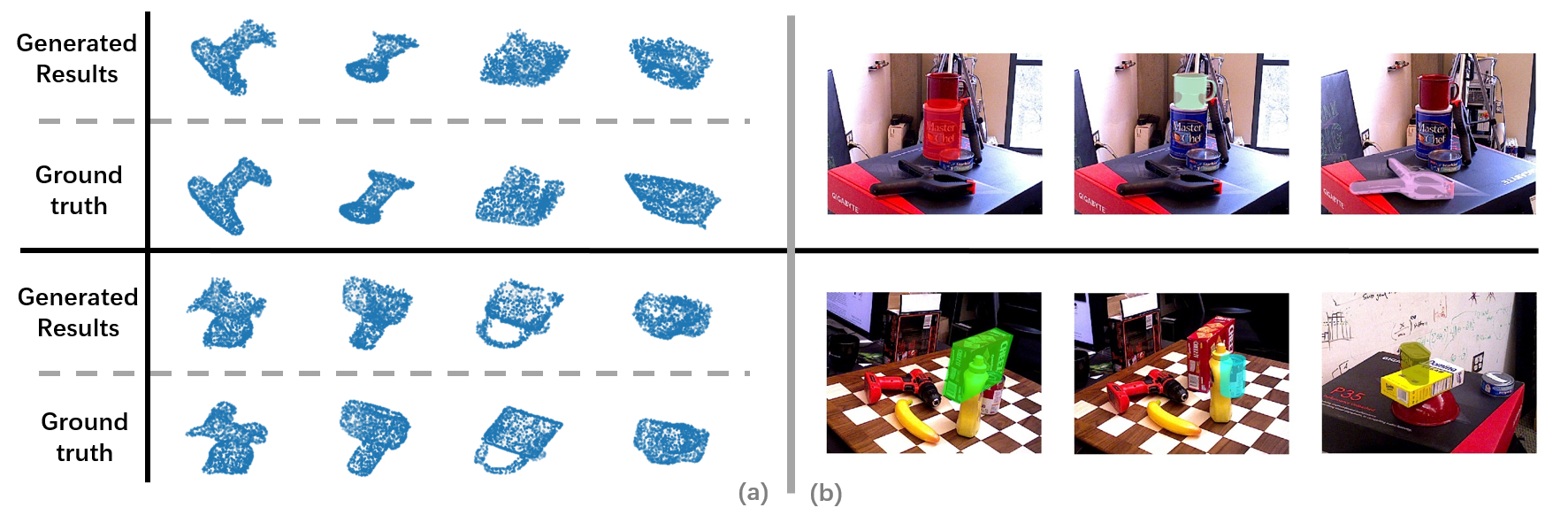}
\end{center}
  % \vspace{-10pt}
  \caption{Qualitative results of our framework: (a) Generating results of object models in target pose on LineMOD dataset, and (b) Pose estimation results on YCB-Video dataset.}
  \label{fig_quali1}
  % \vspace{-10pt}
\end{figure*}

\subsection{Learning Objective}
\label{3.2}
We can represent 6D pose by a homogeneous transformation matrix, $p = [R|t] \in SE(3)$. To be specific, a 6D pose is described by a rotation matrix $R \in SO(3)$ and a translation vector $t \in \mathbb{R}^3$. In our experiments, the estimated pose is defined in the camera coordinate system. We use quaternion to represent the 3D rotation and algebraically convert the quaternion into the rotation matrix to calculate the loss.

\textbf{Pose Estimation Loss.}
Inspired by \cite{Hinterstoisser2012ADD}, for a single pair of pose representation $p = [R|t]$, the loss for 6D pose estimation can be defined as \cite{Xiang2018posecnn, Wang2019DenseFusion}
\begin{equation}
    L_{p}(p,\hat{p}) = \frac{1}{M}\sum_j{ || (Rx_j+t) - (\hat{R}x_j+\hat{t}) || }
\end{equation}
where $x_j$ denotes the $j^{th}$ point of the $M$ randomly selected 3D points from the object’s 3D model, $p = [R|t]$ is the target pose, and $\hat{p} = [\hat{R}|\hat{t}]$ is the estimated pose.
Unfortunately, the above loss function does not handle symmetric objects appropriately, since the canonical orientation of a symmetric object is not well-defined, which leads to multiple correct 3D rotations. Using such a loss function on symmetric objects imposes unnecessary constraints on the networks, i.e., regressing to one of the alternative 3D rotations, thereby giving possibly inconsistent training signals. Therefore, for symmetric objects, we instead minimize the offset between each point on the estimated model and the closest point on the ground truth object model. The loss function becomes:
\begin{equation}
     L_{p}(p,\hat{p}) = \frac{1}{M}\sum_j{ \min_{0<k<M}|| (Rx_j+t) - (\hat{R}x_k+\hat{t}) || }
     \label{PLS}
\end{equation}

Since our final goal is to choose the best result among the per-point estimation, we use the point-wise confidence scores to modulate the per-point loss as follows:
\begin{equation}
     L_{P}(p,\hat{p}) = \sum_i{ c_{i}L^{i}_{p}(p,\hat{p}) }
     \label{confidence_pose_loss}
\end{equation}
where $c_{i}$ represents the confidence score corresponding with $i^{th}$ prediction. Intuitively, estimation candidates leading to low pose estimation loss will have higher confidence scores. In the experiments, we view the estimation result that has the highest confidence as the final final. % Note that since the $N$ confidence scores are normalized by softmax activation function, it is safe to exclude the confidence penalty term $wlog{c_i}$ together with the balancing hyperparameter $w$ mentioned in \cite{Wang2019DenseFusion}.

\textbf{Generation Loss.}
Given the set of target points $S$ and the generated point set $\hat{S}$, the generation error for $\hat{S}$ can be computed as the (extended) Chamfer distance \cite{Yang2017FoldingNet}:
\begin{equation}
\begin{aligned}
    L_{CD}(S,\hat{S}) = \max \{
            \frac{1}{|S|}\sum_{ \mathbf{x}\in S }
            {            \min_{\hat{\mathbf{x}}\in\hat {S}}
               {|| \mathbf{x} - \mathbf{\hat{x}} ||_2}
            },\\
            \frac{1}{|\hat{S}|}\sum_{ \mathbf{\hat{x}}\in \hat{S} }
            {
               \min_{\mathbf{x}\in S}
               {|| \mathbf{\hat{x}} - \mathbf{x} ||_2}
            \}
        }
\end{aligned}
\label{CD}
\end{equation}
The term $\min_{\hat{\mathbf{x}}\in\hat{S}}{|| \mathbf{x} - \mathbf{\hat{x}} ||_2}$ enforces that any 3D point $\mathbf{x}$ in the original point cloud is assigned to a matching 3D point $\mathbf{\hat{x}}$ in the generated point cloud, and the term $ \min_{\mathbf{x}\in S}{|| \mathbf{\hat{x}} - \mathbf{x} ||_2}$ enforces the matching vice versa. The max operation enforces simultaneous minimization of the distance from $S$ to $\hat{S}$ and the distance vice versa.

The overall training loss will then be
\begin{equation}
    L = \lambda_1L_{P}(p,\hat{p}) + \lambda_2L_{CD}(S_{cano},\hat{S}_{cano}) + \lambda_3L_{CD}(S_{gt},\hat{S}_{gt})
\end{equation}
where $S_{cano}$, $S_{gt}, \hat{S}_{cano}$ and $\hat{S}_{gt}$ denote the real point cloud in canonical pose, the real point cloud in target pose, the generated point cloud in canonical pose, and the generated point cloud in target pose, respectively. $\lambda_1, \lambda_2$ and $\lambda_3$ are hyper-parameters adjusted to facilitate the training procedure: all of them are set to 1 to jointly train the networks.

% % \vspace{-15pt}
\subsection{Technical Details}
\label{3.3}
\textbf{Training Strategy.}
We employ a multi-step training paradigm to train the model. First, we jointly train the P$^2$GNet to perform pose-guided point cloud generation and pose estimation, which in the mean time incorporates the knowledge of object models into the RGB embedding sub-network contained in the point-wise fusion network. Next, following the training paradigm described in DenseFusion \cite{Wang2019DenseFusion}, we can also integrate the iterative refinement network into our model, and enable iterative refinement to further improve the performance of our model.

\textbf{Implementation Details.}
We use the same CNN-based RGB embedding network as in DenseFusion \cite{Wang2019DenseFusion} which features a PSPNet \cite{Zhao2017PsPNet} using ResNet18 \cite{He2016Residue} as the backbone. Noticing that DenseFusion \cite{Wang2019DenseFusion} didn't use BatchNorm in its RGB embedding network, we conduct ablative experiments (see Sec.~\ref{4.3}) to demonstrate the effectiveness of our method. Besides, due to the limitation of our GPU resources, we use only 1 refinement iteration and sample $512$ points for all experiments as opposed to 2 iterations and $1000$ points used in DenseFusion, which gives the baseline method an extra advantage.

% % \vspace{-0.2pt}
\section{Experiments}
% \vspace{-2pt}
In this section, we present both qualitative and quantitative results of our framework. On LineMOD \cite{Hinterstoisser2012ADD} dataset, we compare our method with state-of-the-art methods using geometric template/feature-based techniques \cite{Brachmann2014L6D, Brachmann2016Uncertainty}, as well as recently emerged learning-based methods \cite{Kehl2017SSD6D, Xiang2018posecnn, Sundermeyer2018Implicit, Wang2019DenseFusion}. We also evaluate our method on YCB-Video dataset \cite{Xiang2018posecnn} to show that our method outperforms the SOTA method, and achieves marked robustness under heavy occlusion.

\subsection{Evaluation Metrics}
% \vspace{-2pt}
We use two metrics to report the performance on the LineMOD dataset and YCB-Video dataset. We use the Average Distance of Model Points (ADD) for non-symmetric objects and ADD-S for the two symmetric objects (eggbox and glue) following prior works \cite{Hinterstoisser2012ADD, Sundermeyer2018Implicit, Tekin2017RealTime}. As for YCB-Video dataset, we report the accuracy in terms of both ADD-S$<$2cm and AUC metrics as in prior works \cite{Wang2019DenseFusion, Xiang2018posecnn}.

\subsection{Evaluation on LineMOD dataset}
% \vspace{-2pt}
\label{4.3}

\begin{table*}
  \caption{Quantitative evaluation of 6D pose estimation (ADD \cite{Hinterstoisser2012ADD}) on the LineMOD dataset. We present four groups of methods: RGB (PoseCNN with DeepIM \cite{Yi2018DeepIM}), RGB-D (Implicit \cite{Sundermeyer2018Implicit} and SSD-6D \cite{Kehl2017SSD6D} using ICP), models w/o and w/ iterative refinement (DenseFusion and ours). SOTA methods \cite{Brachmann2014L6D} (98.3\%) and \cite{Brachmann2016Uncertainty} (99.0\%) are not listed since they didn't report accuracies on each object. Objects with * are symmetric.}
  \label{linemod results 1}
  \centering
  \begin{tabular}{c|cccccccccccccc}
    \hline
        Object & ape & ben. & cam. & can & cat & dri. & duc. & box.*
              & glu.* & hol. & iro. & lam. & pho. & Avg. \\
    \hline
    \tabincell{c}{PCNN w/ DIM}  & 77.0 & 97.5 & 93.5 & 96.5 & 82.1 & 95.0 & 77.7 & 97.1 & 99.4 & 52.8 & 98.3 & 97.5 & 87.7 & 88.6\\
    \hline
    \tabincell{c}{Imp w/ ICP} & 20.6 & 64.3 & 63.2 & 76.1 & 72.0 & 41.6 & 32.4 & 98.6 & 96.4 & 49.9 & 63.1 & 91.7 & 71.0 & 64.7\\
    \tabincell{c}{SSD w/ ICP} & 65 & 80 & 78 & 86 & 70 & 73 & 66 & 100 & 100 & 49 & 78 & 73 & 79 & 79\\
    \hline
    \tabincell{c}{DF w/o IR} & 79.5 & 84.2 & 76.5 & 86.6 & 88.8 & 77.7 & 76.3 & 99.9 & 99.4 & 79.0 & 92.1 & 92.3 & 88.0 & 86.2\\
    \tabincell{c}{Ours w/o IR} & \textbf{92.2} & \textbf{97.1} & \textbf{92.1} & \textbf{96.8} & \textbf{98.8} & \textbf{95.7} & \textbf{89.4} & \textbf{100.0} & 99.4 & \textbf{96.3} & \textbf{98.8} & \textbf{99.0} & \textbf{96.0} & \textbf{96.2}\\
    \hline
    \tabincell{c}{DF w/ IR} & 92.3 & 93.2 & 94.4 & 93.1 & 96.5 & 87.0 & 92.3 & 99.8 & 100.0 & 92.1 & 97.0 & 95.3 & 92.8 & 94.3\\
    \tabincell{c}{Ours w/ IR} & \textbf{92.9} & \textbf{98.2} & \textbf{97.0} & \textbf{97.4} & \textbf{98.1} & \textbf{97.0} & \textbf{95.2} & \textbf{100.0} & 100.0 & \textbf{97.9} & \textbf{98.2} & \textbf{97.7} & \textbf{96.7} & \textbf{97.4}\\
    \hline
  \end{tabular}
  % \vspace{-10pt}
\end{table*}

\textbf{LineMOD Dataset.}
The LineMOD dataset \cite{Hinterstoisser2012ADD} consists of 13 registered video sequences of 13 texture-poor/texture-less 3D objects. It is widely adopted by both feature/template-based methods \cite{Brachmann2014L6D, Brachmann2016Uncertainty, Hinterstoisser2016PPF} and recent learning-based approaches \cite{Yi2018DeepIM, Sundermeyer2018Implicit, Tekin2017RealTime}. We use the same training and testing set as prior learning-based works \cite{Yi2018DeepIM, Rad2017BB8, Sundermeyer2018Implicit} without additional synthetic data.

\begin{table}
  \caption{Ablative study on the effect of \emph{estimation-by-generation paradigm} (G), \emph{point-fusion block} (PF) and \emph{BatchNorm} (BN) on the LineMOD dataset using ADD/ADD-S metric. Objects with * are symmetric.}
  % \vspace{-10pt}
  % \vspace{-2pt}
  \label{linemod results 2}
  \begin{center}
  \begin{tabular}{c|ccc|c}
    \hline
    % \multicolumn{1}{c|}{ } & \multicolumn{1}{|c|}{RGB} & \multicolumn{2}{|c|}{RGB-D} & \multicolumn{2}{|c|}{w/o Iterative Refinement} & \multicolumn{2}{|c}{w/ Iterative Refinement} \\
    % \cmidrule(lr){3-4}\cmidrule(lr){5-6}\cmidrule(lr){7-8}
    % \hline
        Object & \tabincell{c}{Ours w/o \\BN\&PF\&G}
        &  \tabincell{c}{Ours w/o \\BN\&PF}
        &  \tabincell{c}{Ours w/o \\BN}
        &  Ours \\
    \hline
    ape       & 69.3  & 81.7 & 89.7 & 92.2  \\
    bench v.  & 83.9  & 92.1 & 93.5 & 97.1  \\
    camera    & 68.5  & 83.7 & 92.1 & 92.1  \\
    can       & 80.5  & 88.3 & 94.2 & 96.8  \\
    cat       & 89.2  & 92.4 & 92.7 & 98.8  \\
    driller   & 75.4  & 85.2 & 90.7 & 95.7  \\
    duck      & 57.5  & 80.3 & 89.4 & 89.4  \\
    eggbox*   & 100.0 & 99.9 & 99.9 & 100.0 \\
    glue*     & 99.6  & 99.9 & 99.9 & 99.4 \\
    hole p.   & 63.2  & 85.9 & 87.3 & 96.3  \\
    iron      & 90.3  & 94.8 & 95.6 & 98.8  \\
    lamp      & 91.6  & 96.4 & 97.9 & 99.0  \\
    phone     & 88.3  & 93.0 & 95.3 & 96.0  \\
    \hline
    MEAN      & 81.3  & 90.3 & 93.7 & 96.2  \\
    \hline
  \end{tabular}
  \end{center}
  % \vspace{-27pt}
\end{table}

We compared our method with previous RGB methods with depth refinement (ICP) \cite{Sundermeyer2018Implicit, Kehl2017SSD6D} and RGB-D fusion method \cite{Wang2019DenseFusion} on the ADD/ADD-S metric as presented in Table~\ref{linemod results 1}. The results of the color-based state-of-the-art method \cite{Xiang2018posecnn, Yi2018DeepIM} is also listed for reference. Without iterative refinement step, our method outperforms 10.0\% over the baseline method and 1.9\% over the baseline method with iterative refinement, proving that the proposed estimation-by-generation scheme helps to achieve accurate pose estimation even without post-processing steps. With iterative refinement, to our knowledge, P$^2$GNet achieves state-of-the-art performance (97.4\%) in all single-view RGB-D input network-based methods, approaching the overall SOTA performance: \cite{Brachmann2014L6D} reports 98.3\% and \cite{Brachmann2016Uncertainty} reports 99.0\% on this dataset.  Strictly speaking, however, these methods are not directly comparable, since they are based on hand-crafted feature/template and take $\sim$0.5s to process a single instance, which is approximately 10 to 20 times slower than our pipeline, and even 200x slower if we don't take the pre-segmentation step into consideration. As reported in \cite{Brachmann2014L6D}, their method only achieved 96.4\% accuracy when reducing the processing time to $\sim$150ms.

\textbf{Ablative study of the design.}
Table~\ref{linemod results 2} summarizes the ablation studies on critical components of P$^2$GNet\footnote{The improvement induced by squeeze-and-excitation block is trivial (less than 0.5\%), and is therefore not included}. The first column indicates that simply introducing object point cloud degrades the model performance (81.3\% vs. 86.2\%\footnote{Note that our model w/o BatchNorm, point-fusion blocks and pose-guided point cloud generating task is equivalent to DenseFusion with an additional input branch.}), as the object point clouds are not exactly aligned with their corresponding point from RGB-D images. In addition, we observe intensified oscillation of the training loss when conducting the experiments, which confirms our assumption that unaligned point cloud input can be a potential perturbation to the network. Comparing the first and the second column shows the effectiveness of our estimation-by-generation scheme as performing pose-guided point cloud task provides the networks with very significant improvements. Employing point-fusion blocks brings an extra 3.4\% improvement in performance, which demonstrates that point-fusion block can partly solve the matching problem between point clouds and pixels. For qualitative analysis, we visualize the generated point clouds in Figure~\ref{fig_quali1}a. The performance of P$^2$GNet on point cloud generating task indicates its capability of extracting intrinsic properties of object models. % Our model takes the object models in canonical poses as input and produces object models in target pose.

\subsection{Evaluation on YCB-Video Dataset}
% \vspace{-2pt}
\label{4.4}
\begin{table*}[!htbp]
  \caption{Quantitative evaluation of 6D pose (ADD-S \cite{Hinterstoisser2012ADD} and AUC \cite{Xiang2018posecnn}) on the YCB-Video dataset. Objects with * are symmetric. IR is the abbreviation of Iterative Refinement.}
  \label{ycb results 1}
  \centering
  \begin{tabular}{c|cc|cccc|cccc|c}
    \hline
    \multicolumn{1}{c|}{ } & \multicolumn{2}{|c|}{PoseCNN+ICP} & \multicolumn{2}{|c}{DenseFusion} & \multicolumn{2}{c|}{Ours} &
    \multicolumn{2}{|c}{DenseFusion+IR}        &
    \multicolumn{2}{c|}{Ours+IR}        &
    \multicolumn{1}{|c}{MV5-MCN}\\
    \hline
        Object &  AUC
               &  $<$2cm
               &  AUC
               &  $<$2cm
               &  AUC
               &  $<$2cm
               &  AUC
               &  $<$2cm
               &  AUC
               &  $<$2cm
               &  $<$2cm \\
    \hline
    002\_chf\_can    & 95.8 & 100.0 & \textbf{95.2} & 100.0 & 93.7 & 100.0 & \textbf{96.4} & 100.0 & 92.3 & 100.0 & 96.2\\
    003\_ckr\_box    & 92.7 & 91.6  & \textbf{92.5} & 99.3  & 92.1 & \textbf{99.6}  & \textbf{95.5} & 99.5  & 92.9 & \textbf{100.0} & 90.9\\
    004\_sgr\_box    & 98.2 & 100.0 & 95.1 & 100.0 & \textbf{95.4} & 100.0 & \textbf{97.5} & 100.0 & 96.4 & 100.0 & 95.3\\
    005\_sop\_can    & 94.5 & 96.9  & 93.7 & 96.9  & \textbf{95.9} & \textbf{100.0} & 94.6 & 96.9  & \textbf{95.6} & \textbf{100.0} & 97.5\\
    006\_mtd\_bottle & 98.6 & 100.0 & \textbf{95.9} & 100.0 & 95.7 & 100.0 & \textbf{97.2} & 100.0 & 95.7 & 100.0 & 97.0\\
    007\_fish\_can   & 97.1 & 100.0 & \textbf{94.9} & 100.0 & 94.8 & 100.0 & \textbf{96.6} & 100.0 & 95.3 & 100.0 & 95.1\\
    008\_pud\_box    & 97.9 & 100.0 & \textbf{94.7} & \textbf{100.0} & 92.1 & 99.1 & \textbf{96.5} & 100.0 & 93.5 & 100.0 & 94.5\\
    009\_glt\_box    & 98.8 & 100.0 & 95.8 & 100.0 & \textbf{96.1} & 100.0 & \textbf{98.1} & 100.0 & 96.8 & 100.0 & 96.0\\
    010\_meat\_can   & 92.7 & 93.6  & 90.1 & 93.1  & \textbf{95.6} & \textbf{98.7} & 91.3 & 93.1  & \textbf{95.7} & \textbf{98.9} & 96.7\\
    011\_banana      & 97.1 & 99.7  & 91.5 & 93.9  & \textbf{93.7} & \textbf{98.7}  & \textbf{96.6} & 100.0 & 94.2 & 100.0 & 94.4\\
    019\_pit\_base    & 97.8 & 100.0 & \textbf{94.6} & 100.0 & 93.6 & 100.0 & \textbf{97.1} & 100.0 & 94.1 & 100.0 & 96.2\\
    021\_cleanser    & 96.9 & 99.4  & 94.3 & \textbf{99.8} & \textbf{94.4} & 99.7 & \textbf{95.8} & 100.0 & 94.7 & 100.0 & 95.4\\
    024\_bowl*       & 81.0 & 54.9  & 86.6 & 69.5 & \textbf{88.0} & \textbf{75.6} & 88.2 & 98.8  & \textbf{89.6} & \textbf{100.0} & 82.0\\
    025\_mug         & 95.0 & 99.8  & \textbf{95.5} & 100.0 & 95.2 & 100.0 & \textbf{97.1} & 100.0 & 95.1 & 100.0 & 96.8\\
    035\_drill       & 98.2 & 99.6  & 92.4 & 97.1  & \textbf{93.5} & \textbf{100.0} & \textbf{96.0} & 98.7 & 95.3 & \textbf{100.0} & 93.1\\
    036\_block*      & 87.6 & 80.2  & 85.5 & \textbf{93.4} & \textbf{86.0} & 87.7 & \textbf{89.7} & 94.6 & 85.6 & \textbf{100.0} & 93.6\\
    037\_scissors    & 91.7 & 95.6  & \textbf{96.4} & \textbf{100.0} & 88.0 & 89.0  & \textbf{95.2} & 100.0 & 90.3 & 100.0 & 94.2\\
    040\_marker      & 97.2 & 99.7  & \textbf{94.7} & 99.2  & 94.1 & \textbf{99.5}  & \textbf{97.5} & 100.0 & 95.4 & 100.0 & 95.4\\
    051\_clamp*      & 75.2 & 74.9  & 71.6 & 78.5  & \textbf{88.7} & \textbf{94.3}  & 72.9 & 79.2  & \textbf{90.8} & \textbf{99.2} & 93.3\\
    052\_ex\_clamp*   & 64.4 & 48.8  & 69.0 & 69.5  & \textbf{83.7} & \textbf{73.7}  & 69.8 & 76.3  & \textbf{89.1} & \textbf{99.6} & 90.9\\
    061\_brick*      & 97.2 & 100.0 & 92.4 & 100.0 & \textbf{92.8} & 100.0 & 92.5 & 100.0 & \textbf{94.6} & 100.0 & 95.9\\
    \hline
    MEAN             & 93.0 & 93.2  & 91.2 & 95.3  & \textbf{93.1} & \textbf{97.5} & 93.1 & 96.8 & \textbf{93.9} & \textbf{99.9} & 94.3\\
    \hline
  \end{tabular}
  % \vspace{-15pt}
\end{table*}

\textbf{YCB-Video Dataset.}
The YCB-Video Dataset \cite{Xiang2018posecnn} features 21 YCB objects of varying shapes and texture levels under different occlusion conditions. The dataset contains 92 RGB-D videos, where each video shows a subset of the 21 objects in different indoor scenes. To ensure a fair comparison, we follow prior work \cite{Wang2019DenseFusion} using the same 80 videos and 80,000 synthetic images released by \cite{Xiang2018posecnn} for training, and 2,949 key frames chosen from the rest 12 videos for testing. During the testing procedure, we use the same segmentation masks as in PoseCNN.

Table~\ref{ycb results 1} demonstrates that our method without iterative consistently surpasses DenseFusion with or without iterative refinement. With iterative refinement, P$^2$GNet achieves 99.9\% accuracy in terms of ADD-S$<$2cm metric, and is comparable with methods using multi-view information (93.9\% vs.94.3\%). Still, we have to clarify that all the methods except MV5-MCN \cite{Li2018MCN} listed in table~\ref{ycb results 1} use the same segmentation masks released by PoseCNN, which induces a performance drop because of the poor detection result. Using masks provided in YCB-Video dataset, our method achieves 97.8\% (ADD-S$<$2cm), 93.3\% (AUC) without iterative refinement, and 94.2\% (AUC) with iterative refinement. By comparing with the state-of-the-art methods on YCB-Video dataset, we summarize that the proposed estimation-by-generation scheme has following advantages: a) instance-aware estimation: one obstacle for achieving better performance on the YCB-Video dataset is to find discriminative features other than object appearance for the object \emph{051\_clamp} and \emph{052\_ex\_clamp}, since their appearances are largely the same. Our method provides a possible solution by learning object 3D information, resulting in $\sim$15\% improvement in each object; b) addressing single-view appearance ambiguity: our method uses only single-view input and object point cloud, while achieving performance comparable to methods using multi-view information. We suggest that our method is more practical in application since circumstances where only single-view inputs are provided are more common.

\textbf{Robustness towards heavy occlusion.}
Figure~\ref{fig_quali1}b displays some 6D pose estimation results on the YCB-Video dataset. We can see that the prediction is quite accurate even if the center is occluded by another object, indicating the robustness of P$^2$GNet towards heavy occlusion. We attribute this to the proposed pose-guided point cloud generating task which enforces the network to capture the appearance and geometric information of the object models well even under adversarial circumstances.
% \begin{figure}[htbp]
% \begin{center}
% % \includegraphics[width=\textwidth]{quali3.png}
% \end{center}
%   \caption{}
%   \label{fig_quali2}
%   % \vspace{-10pt}
% \end{figure}
\subsection{Time Efficiency}
% \vspace{-2pt}
Given a $240 \times 320$ pre-segmented image, P$^2$GNet model takes $\sim$1.4ms to process a single object instance, and $\sim$1.1ms to run iterative refinement on the instance, using batch parallelization on a desktop computer with an Intel i7 3.7GHz CPU and a GTX 1080 Ti GPU.  Our implementation for SegNet \cite{Xiang2018posecnn} takes $\sim$29.8 ms for segmentation, becoming the bottleneck of system efficiency. Overall, the inference speed is fast enough for real-time application ($\sim$25 FPS, about 5 objects each frame).

\section{Conclusion and Future Works}
% \vspace{-2pt}
We present a novel approach to estimate 6D poses of known objects. Our approach explores how to integrate the intrinsic properties of object models into the state-of-the-art single-view-based systems. With the proposed estimation-by-generation scheme, our method outperforms previous approaches on two widely used datasets, approaching the performance of multi-view-based methods, while achieving real-time inference. We hope our system will inspire future research along this challenging but rewarding research direction. Specifically, it would be interesting to explore the limits of the approach w.r.t. the number of objects. We also suggest developing voting-based estimation scheme on top of the system to achieve better performance.

% \newpage
{\small
\bibliographystyle{ieee_fullname}
\bibliography{Ref}

\begin{thebibliography}{10}\itemsep=-1pt

\bibitem{achlioptas20173dad}
Panos Achlioptas, Olga Diamanti, Ioannis Mitliagkas, and Leonidas Guibas.
\newblock Learning representations and generative models for 3d point clouds.
\newblock In {\em ICLR}, 2018.

\bibitem{Brachmann2014L6D}
Eric Brachmann, Alexander Krull, Frank Michel, Stefan Gumhold, Jamie Shotton,
  and Carsten Rother.
\newblock Learning 6d object pose estimation using 3d object coordinates.
\newblock In {\em ECCV}, 2014.

\bibitem{Brachmann2016Uncertainty}
Eric Brachmann, Frank Michel, Alexander Krull, Michael~Ying Yang, Stefan
  Gumhold, and Carsten Rother.
\newblock Uncertainty-driven 6d pose estimation of objects and scenes from a
  single rgb image.
\newblock In {\em CVPR}, 2016.

\bibitem{Brock2016GDVM}
Andrew Brock, T Lim, James Ritchie, and Nick Weston.
\newblock Generative and discriminative voxel modeling with convolutional
  neural networks.
\newblock In {\em NeurIPS}, 2016.

\bibitem{Buch2017Rotational}
Anders~Glent Buch, Lilita Kiforenko, Dirk Kraft, Anders~Glent Buch, Lilita
  Kiforenko, and Dirk Kraft.
\newblock Rotational subgroup voting and pose clustering for robust 3d object
  recognition.
\newblock In {\em ICCV}, 2017.

\bibitem{Chen2016multiview3d}
Xiaozhi Chen, Huimin Ma, Ji Wan, Bo Li, and Tian Xia.
\newblock Multi-view 3d object detection network for autonomous driving.
\newblock In {\em CVPR}, 2017.

\bibitem{Collet2011MOPED}
Alvaro Collet, Manuel Martinez, and Siddhartha Srinivasa.
\newblock The moped framework: Object recognition and pose estimation for
  manipulation.
\newblock {\em I. J. Robotic Res.}, Sep. 2011.

\bibitem{Dai2017scannet}
Angela Dai, Angel Chang, Manolis Savva, Maciej Halber, Thomas Funkhouser, and
  Matthias Nießner.
\newblock Scannet: Richly-annotated 3d reconstructions of indoor scenes.
\newblock In {\em CVPR}, 2017.

\bibitem{Fisher01projectiveicp}
Robert~B. Fisher.
\newblock Projective icp and stabilizing architectural augmented reality
  overlays.
\newblock In {\em Proc. Int. Symp. on Virtual and Augmented Architecture
  (VAA)}, 2001.

\bibitem{Geiger2012KITTI}
A. {Geiger}, P. {Lenz}, and R. {Urtasun}.
\newblock Are we ready for autonomous driving? the kitti vision benchmark
  suite.
\newblock In {\em CVPR}, 2012.

\bibitem{He2016Residue}
Kaiming He, Xiangyu Zhang, Shaoqing Ren, and Sun Jian.
\newblock Deep residual learning for image recognition.
\newblock In {\em CVPR}, 2016.

\bibitem{Hinterstoisser2011linemod}
Stefan Hinterstoisser, Stefan Holzer, Cedric Cagniart, Slobodan Ilic, Kurt
  Konolige, Nassir Navab, and Vincent Lepetit.
\newblock Multimodal templates for real-time detection of texture-less objects
  in heavily cluttered scenes.
\newblock In {\em ICCV}, 2011.

\bibitem{Hinterstoisser2012ADD}
Stefan Hinterstoisser, Vincent Lepetit, Slobodan Ilic, Stefan Holzer, Gary
  Bradski, Kurt Konolige, and Nassir Navab.
\newblock Model based training, detection and pose estimation of texture-less
  3d objects in heavily cluttered scenes.
\newblock In {\em ACCV}, 2012.

\bibitem{Hinterstoisser2016PPF}
Stefan Hinterstoisser, Vincent Lepetit, Naresh Rajkumar, and Kurt Konolige.
\newblock Going further with point pair features.
\newblock In {\em ECCV}, 2016.

\bibitem{Hodan2018BOP}
Tomas Hodan, Frank Michel, Eric Brachmann, Wadim Kehl, Anders GlentBuch, Dirk
  Kraft, Bertram Drost, Joel Vidal, Stephan Ihrke, Xenophon Zabulis, Caner
  Sahin, Fabian Manhardt, Federico Tombari, Tae-Kyun Kim, Jiri Matas, and
  Carsten Rother.
\newblock Bop: Benchmark for 6d object pose estimation.
\newblock In {\em ECCV}, 2018.

\bibitem{Hu2017Se}
Jie Hu, Li Shen, and Gang Sun.
\newblock Squeeze-and-excitation networks.
\newblock {\em TPAMI}, Sep. 2017.

\bibitem{Kehl2017SSD6D}
W. {Kehl}, F. {Manhardt}, F. {Tombari}, S. {Ilic}, and N. {Navab}.
\newblock Ssd-6d: Making rgb-based 3d detection and 6d pose estimation great
  again.
\newblock In {\em ICCV}, 2017.

\bibitem{Kehl2016Local}
Wadim Kehl, Fausto Milletari, Federico Tombari, Slobodan Ilic, and Nassir
  Navab.
\newblock Deep learning of local rgb-d patches for 3d object detection and 6d
  pose estimation.
\newblock In {\em ECCV}, 2016.

\bibitem{klokov2017kdnet}
Roman Klokov and Victor Lempitsky.
\newblock Escape from cells: Deep kd-networks for the recognition of 3d point
  cloud models.
\newblock In {\em ICCV}, 2017.

\bibitem{Li2018MCN}
Chi Li, Jin Bai, and Gregory Hager.
\newblock A unified framework for multi-view multi-class object pose
  estimation.
\newblock In {\em ECCV}, 2018.

\bibitem{Marchand2016AR}
E. {Marchand}, H. {Uchiyama}, and F. {Spindler}.
\newblock Pose estimation for augmented reality: A hands-on survey.
\newblock {\em TVCG}, Dec 2016.

\bibitem{Rad2017BB8}
Mahdi Rad and Vincent Lepetit.
\newblock Bb8: A scalable, accurate, robust to partial occlusion method for
  predicting the 3d poses of challenging objects without using depth.
\newblock In {\em ICCV}, 2017.

\bibitem{Riegler2017octnet}
Gernot Riegler, Ali Ulusoy, and Andreas Geiger.
\newblock Octnet: Learning deep 3d representations at high resolutions.
\newblock In {\em CVPR}, 2017.

\bibitem{Rios2013Discriminatively}
Reyes Rios-Cabrera and Tinne Tuytelaars.
\newblock Discriminatively trained templates for 3d object detection: A real
  time scalable approach.
\newblock In {\em ICCV}, 2013.

\bibitem{Qi2018frustum}
Charles Ruizhongtai~Qi, Wei Liu, Chenxia Wu, Hao Su, and Leonidas Guibas.
\newblock Frustum pointnets for 3d object detection from rgb-d data.
\newblock In {\em CVPR}, 2018.

\bibitem{Qi2016PointNet}
Charles Ruizhongtai~Qi, Hao Su, Kaichun Mo, and Leonidas J.~Guibas.
\newblock Pointnet: Deep learning on point sets for 3d classification and
  segmentation.
\newblock In {\em CVPR}, 2016.

\bibitem{Qi2017pointnet++}
Charles Ruizhongtai~Qi, Li Yi, Hao Su, and Leonidas Guibas.
\newblock Pointnet++: Deep hierarchical feature learning on point sets in a
  metric space.
\newblock In {\em NeurIPS}, 2017.

\bibitem{Schwarz2015RGB}
Max Schwarz, Hannes Schulz, and Sven Behnke.
\newblock Rgb-d object recognition and pose estimation based on pre-trained
  convolutional neural network features.
\newblock In {\em ICRA}, 2015.

\bibitem{Sundermeyer2018Implicit}
Martin Sundermeyer, Zoltan~Csaba Marton, Maximilian Durner, Manuel Brucker, and
  Rudolph Triebel.
\newblock Implicit 3d orientation learning for 6d object detection from rgb
  images.
\newblock In {\em ECCV}, 2018.

\bibitem{Tejani2014Latent}
Alykhan Tejani, Danhang Tang, Rigas Kouskouridas, and Tae~Kyun Kim.
\newblock Latent-class hough forests for 3d object detection and pose
  estimation.
\newblock In {\em ECCV}, 2014.

\bibitem{Tekin2017RealTime}
Bugra Tekin, Sudipta~N. Sinha, and Pascal Fua.
\newblock Real-time seamless single shot 6d object pose prediction.
\newblock In {\em CVPR}, 2018.

\bibitem{Tremblay2018Deep}
Jonathan Tremblay, Thang To, Balakumar Sundaralingam, Yu Xiang, Dieter Fox, and
  Stan Birchfield.
\newblock Deep object pose estimation for semantic robotic grasping of
  household objects.
\newblock In {\em CoRL}, 2018.

\bibitem{Wang2019DenseFusion}
Chen Wang, Danfei Xu, Yuke Zhu, Roberto Martín-Martín, and Silvio Savarese.
\newblock Densefusion: 6d object pose estimation by iterative dense fusion.
\newblock In {\em CVPR}, 2019.

\bibitem{Wang2018NonLocal}
X. {Wang}, R. {Girshick}, A. {Gupta}, and K. {He}.
\newblock Non-local neural networks.
\newblock In {\em CVPR}, 2018.

\bibitem{Wohlhart2015Descriptors}
P. {Wohlhart} and V. {Lepetit}.
\newblock Learning descriptors for object recognition and 3d pose estimation.
\newblock In {\em CVPR}, 2015.

\bibitem{Wu20163DGAN}
Jiajun Wu, Chengkai Zhang, Tianfan Xue, William T.~Freeman, and Joshua
  B.~Tenenbaum.
\newblock Learning a probabilistic latent space of object shapes via 3d
  generative-adversarial modeling.
\newblock In {\em NeurIPS}, 2016.

\bibitem{Xiang2018posecnn}
Yu Xiang, Tanner Schmidt, Venkatraman Narayanan, and Dieter Fox.
\newblock Posecnn: A convolutional neural network for 6d object pose estimation
  in cluttered scenes.
\newblock In {\em RSS}, 2018.

\bibitem{Xu2018Pointfusion}
D. {Xu}, D. {Anguelov}, and A. {Jain}.
\newblock Pointfusion: Deep sensor fusion for 3d bounding box estimation.
\newblock In {\em CVPR}, 2018.

\bibitem{Yang2017FoldingNet}
Yaoqing Yang, Chen Feng, Yiru Shen, and Dong Tian.
\newblock Foldingnet: Point cloud auto-encoder via deep grid deformation.
\newblock In {\em CVPR}, 2018.

\bibitem{Yi2018DeepIM}
Li Yi, Wang Gu, Xiangyang Ji, Xiang Yu, and Dieter Fox.
\newblock Deepim: Deep iterative matching for 6d pose estimation.
\newblock In {\em ECCV}, 2018.

\bibitem{Zhao2017PsPNet}
H. {Zhao}, J. {Shi}, X. {Qi}, X. {Wang}, and J. {Jia}.
\newblock Pyramid scene parsing network.
\newblock In {\em CVPR}, 2017.

\bibitem{Zhou2017voxelnet}
Yin Zhou and Oncel Tuzel.
\newblock Voxelnet: End-to-end learning for point cloud based 3d object
  detection.
\newblock In {\em CVPR}, 2018.

\bibitem{Zhu2014Single}
M. {Zhu}, K.~G. {Derpanis}, Y. {Yang}, S. {Brahmbhatt}, M. {Zhang}, C.
  {Phillips}, M. {Lecce}, and K. {Daniilidis}.
\newblock Single image 3d object detection and pose estimation for grasping.
\newblock In {\em ICRA}, May 2014.

\end{thebibliography}
}

\end{document}